\documentclass{article} 
\usepackage[final]{nips_2017}
\usepackage{svg}
\usepackage{amssymb,amsmath}
\usepackage{ifxetex,ifluatex}
\usepackage{fixltx2e} 
\usepackage{longtable,booktabs}
\usepackage{floatrow}
\DeclareMathOperator*{\expect}{\scalerel*{\mathbb{E}}{\textstyle\sum}}

\DeclareMathOperator*{\argmax}{arg\,max}
\usepackage{scalerel}

\ifnum 0\ifxetex 1\fi\ifluatex 1\fi=0 
  \usepackage[T1]{fontenc}
  \usepackage[utf8]{inputenc}
\else 
  \ifxetex
    \usepackage{mathspec}
    \usepackage{xltxtra,xunicode}
  \else
    \usepackage{fontspec}
  \fi
  \defaultfontfeatures{Mapping=tex-text,Scale=MatchLowercase}
  
\fi
\IfFileExists{upquote.sty}{\usepackage{upquote}}{}
\ifxetex
  \usepackage[setpagesize=false, 
              unicode=false, 
              xetex]{hyperref}
\else
  \usepackage[unicode=true]{hyperref}
\fi

\usepackage{letltxmacro}
\LetLtxMacro{\oldhypertarget}{\hypertarget}
\renewcommand{\hypertarget}[2]{\leavevmode\oldhypertarget{#1}{#2}}

\hypersetup{breaklinks=true,
            bookmarks=true,
            pdfauthor={},
            pdftitle={Disentangling Video with Independent Prediction},
            colorlinks=true,
            citecolor=blue,
            urlcolor=blue,
            linkcolor=magenta,
            pdfborder={0 0 0}}
\usepackage{fancyvrb}

\DefineVerbatimEnvironment{Highlighting}{Verbatim}{commandchars=\\\{\}}
\newenvironment{Shaded}{}{}

\newcommand{\DecValTok}[1]{\textcolor[rgb]{0.25,0.63,0.44}{#1}}

\newcommand{\FloatTok}[1]{\textcolor[rgb]{0.25,0.63,0.44}{#1}}

\newcommand{\VariableTok}[1]{\textcolor[rgb]{0.10,0.09,0.49}{#1}}

\newcommand{\OperatorTok}[1]{\textcolor[rgb]{0.40,0.40,0.40}{#1}}

\newcommand{\NormalTok}[1]{#1}
\usepackage{graphicx,grffile}
\makeatletter
\def\maxwidth{\ifdim\Gin@nat@width>\linewidth\linewidth\else\Gin@nat@width\fi}
\def\maxheight{\ifdim\Gin@nat@height>\textheight\textheight\else\Gin@nat@height\fi}
\makeatother
\setkeys{Gin}{width=\maxwidth,height=\maxheight,keepaspectratio}
\usepackage{amsfonts}       
\usepackage{nicefrac}       
\usepackage{microtype}      

\title{Disentangling Video with Independent Prediction}
    \author{
                                        William F. Whitney
                                                            and Rob Fergus\\
                Department of Computer Science\\
                New York University\\
                \texttt{\{wwhitney,fergus\}@cs.nyu.edu}
                            }
\date{}

\usepackage{subfig}
\AtBeginDocument{%

}
\AtBeginDocument{%

}
\usepackage{float}
\floatstyle{ruled}
\makeatletter
\@ifundefined{c@chapter}{\newfloat{codelisting}{h}{lop}}{\newfloat{codelisting}{h}{lop}[chapter]}
\makeatother
\floatname{codelisting}{Listing}

\ifx\paragraph\undefined\else
\let\oldparagraph\paragraph
\renewcommand{\paragraph}[1]{\oldparagraph{#1}\mbox{}}
\fi
\ifx\subparagraph\undefined\else
\let\oldsubparagraph\subparagraph
\renewcommand{\subparagraph}[1]{\oldsubparagraph{#1}\mbox{}}
\fi

\begin{document}
\maketitle
\begin{abstract}
We propose an unsupervised variational model for disentangling video
into independent factors, i.e.~each factor's future can be predicted
from its past without considering the others. We show that our approach
often learns factors which are interpretable as objects in a scene.
\end{abstract}

\section{Introduction}\label{introduction}

Deep neural networks have delivered impressive performance on a range of
perceptual tasks, but their distributed representation is difficult to
interpret and poses a challenge for problems involving reasoning.
Motivated by this, the deep learning community has recently explored
methods
{[}\protect\hyperlink{ref-bengio2013representation}{1}--\protect\hyperlink{ref-grathwohl2016disentangling}{4},\protect\hyperlink{ref-Hsu17}{6}--\protect\hyperlink{ref-Janner17}{8},\protect\hyperlink{ref-Greff16}{12},\protect\hyperlink{ref-kulkarni2015deep}{15}--\protect\hyperlink{ref-Siddarth17}{17},\protect\hyperlink{ref-Thomas17}{20},\protect\hyperlink{ref-whitney2016understanding}{21}{]}
for learning distributed representations which are \emph{disentangled},
i.e.~a unit (or small group) within the latent feature vector are
exclusively responsible for capturing distinct concepts within the input
signal. This work proposes such an approach for the video domain, where
the temporal structure of the signal is leveraged to automatically
separate the input into factors that vary independently of one another
over time. Our results demonstrate that these factors correspond to
distinct objects within the video, thus providing a natural
representation for making predictions about future motions of the
objects and subsequent high-level reasoning tasks.

\textbf{Related work.} {[}\protect\hyperlink{ref-wiskott2002slow}{22}{]}
leverage structure at different time-scales to factor signals into
independent components. Our approach can handle multiple factors at the
same time-scale, instead relying on prediction as the factoring
mechanism. Outside of the video domain,
{[}\protect\hyperlink{ref-bengio2013representation}{1}{]} proposed to
disentangle factors that tend to change independently and sparsely in
real-world inputs, while preserving information about them.
{[}\protect\hyperlink{ref-kulkarni2015deep}{15}{]} learned to
disentangle factors of variation in synthetic images using weak
supervision, and
{[}\protect\hyperlink{ref-whitney2016understanding}{21}{]} extended the
method to be fully unsupervised. Similar to our work, both
{[}\protect\hyperlink{ref-Denton17}{3}{]} and
{[}\protect\hyperlink{ref-grathwohl2016disentangling}{4}{]} propose
unsupervised schemes for disentangling video, the latter using a
variational approach. However, ours differs in that it uncovers general
factors, rather than specific ones like identity and pose or
static/dynamic features as these approaches do.

\vspace{-1mm}

\section{Generative model and
inference}\label{generative-model-and-inference}

\begin{figure}
\centering
\includegraphics[width=0.80000\textwidth]{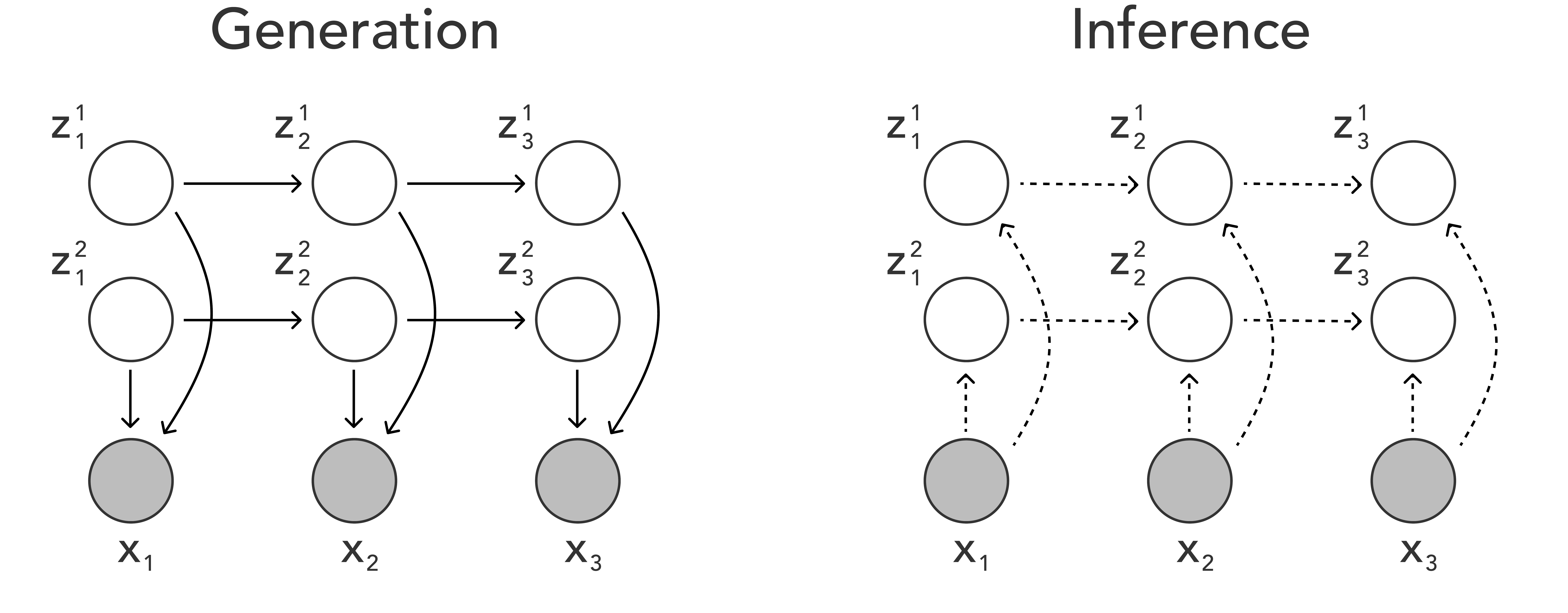}
\caption{\textbf{Left:} Our generative model. Several latent variables
combine to produce each observation, and each variable evolves through
time independently of the others, just like objects which do not
collide. Each observation \(x_t\) is given by a decoder which produces
\(p_\theta(x_t | z^1_{t} ... z^k_t)\). \textbf{Right:} Our variational
inference procedure gives
\(q_\phi(z^1_{t} ... z^k_t | x_t, z^1_{t-1} ... z^k_{t-1}) \approx p(z^1_{t} ... z^k_t | x_t, z^1_{t-1} ... z^k_{t-1})\).}\label{fig:model}
\end{figure}

\vspace{-1mm} The intuition behind our model is simple: in any
real-world scene, most objects do not physically interact with one
another, so their motion can be modeled independently. To find a
representation of videos with these same independences, we introduce an
approach based on a temporal version
{[}\protect\hyperlink{ref-krishnan2017structured}{14}{]} of the
variational auto-encoder
{[}\protect\hyperlink{ref-kingma2013auto}{10}{]}, shown in Figure
\ref{fig:model}. In this model, each video \(X\) comprises a sequence of
frames \(x_1 ... x_n\). \(Z_t = z^1_t...z^k_t\) represents all the
latent factors at time \(t\), where each factor \(z^i_t\) is represented
by a vector. Each of these factors evolve independently from one another
and combine to produce the observation \(x_t\) for timestep \(t\).
Instead of directly maximizing the likelihood \(p(X;\theta)\), we use
variational inference
{[}\protect\hyperlink{ref-hoffman2013stochastic}{5}{]} to optimize the
evidence lower bound (ELBO), that is
\(\theta^*,\phi^* = \argmax_{\theta, \phi}\mathcal{L}(\theta, \phi; X):\)

\begin{equation} \label{eq:elbo}
\expect_{Z \sim q_\phi(Z | X)} \hspace{-5mm}  \log p_\theta(X|Z)  \\
- D_{KL}(q_\phi(Z_1 | X) || p_\theta(Z_1)) \
- \sum_{t=2}^{n} \expect_{\substack{Z_{t-1} \sim \\q(Z_{t-1} | Z_{t-2}, X)} } \
\hspace{-7mm} D_{KL}(q_\phi(Z_t | Z_{t-1}, X) || p_\theta(Z_{t} | Z_{t-1})) \nonumber
\end{equation}

For a derivation, see \protect\hyperlink{sec:elbo-derivation}{Appendix
B}. This lower bound naturally splits into two factors. The first,
\(p_\theta(X|Z)\), is the log-likelihood of the data \(X\) under our
model when sampling from the approximate posterior, that is the
``reconstruction'' of \(X\) from \(Z\). The second factor is the KL
divergence between the learned prior \(p_\theta(Z)\) and the approximate
posterior \(q_\phi(Z|X)\). It represents how far the predictions given
by \(p\) are from the inferred values given by \(q\); it is the
prediction error in the latent space. Our goal is to optimize the space
of Z to make both reconstruction and prediction possible.

For the variational approximation, we choose
\(q_\phi(Z|X) = \prod_{t=1}^n q(Z_t | Z_{t-1}, x_t;\phi)\) which,
analogous to a Kalman filter, considers only the past state and current
observation. This approximation marginalizes over the future, in that
\(q(Z_t | Z_{t-1}, x_t)\) must encode sufficient information to allow
correct predictions of the future as well as fit the prior and the
current observation. This allows us to do inference on a single frame
and ensure our representation retains as much information about the
future as possible. If we wish to use our learned representation for a
downstream task, we may discard the generative model and use the
inference network alone. This inference network can provide a factorized
representation given a single image or use a sequence of images to
produce increasingly tight estimates of the latent variables.

\textbf{Neural network parameterization.} We choose all of the
distributions in our model to be Gaussian with diagonal covariance. To
allow our model to scale to complex nonlinear observations and dynamics,
we parameterize each distribution with a neural network. Table
\ref{tbl:dist-nets} describes each of these parameterizations. As each
distribution is diagonal Gaussian, each network produces two outputs,
corresponding to the mean and the variance of each distribution.

\textbf{Training.} This model can be thought of as a series of
variational autoencoders with a learned prior, and the training
procedure is largely similar to
{[}\protect\hyperlink{ref-kingma2013auto}{10}{]}. At each timestep \(t\)
in a sequence, we compute the prior \(p_\theta(Z_t | Z_{t-1})\) over the
latent space, using \(\mathcal{N}(0, \mathbb{I})\) for \(t=0\). We infer
the approximate posterior \(q_\phi(Z_t | Z_{t-1}, x_t)\) by observing
\(x_t\) and compute the KL divergence \(D_{KL}(q||p)\). We then sample
\(Z_t \sim q_\phi(Z_t | Z_{t-1}, x_t)\) using the reparameterization
trick and compute \(\log p_\theta(x_t | Z_t)\). At the end of a sequence
we update our parameters \(\theta, \phi\) by backprop to maximize the
ELBO (defined above).

\begin{table}[]
\centering
\begin{tabular}[c]{l l l}
\toprule
distribution & parameterization & parameter sharing?\tabularnewline
\midrule
\(p_\theta(z^i_1)\) & \(\mathcal{N}(0, \mathbb{I})\) & N/A (but same for
each \(i\))\tabularnewline
\(p_\theta(z^i_t | z^i_{t-1})\) &
MLP\((d \to 128, 128 \to 128, 128 \to d)\) & across \(t\) (different
params for each \(i\))\tabularnewline
\(p_\theta(x_t | Z_t)\) & DCGAN generator & across \(t\)\tabularnewline
\(q_\phi(Z_t | Z_{t-1}, x_t)\) & DCGAN discriminator & across
\(t\)\tabularnewline
\bottomrule
\end{tabular}

\caption{\label{tbl:dist-nets}The parameterization for each of the
distributions in our model. Each latent factor has its own transition
function, but all modules are shared across all timesteps. The variable
\(d\) is the dimension of the entire latent space including all factors.
DCGAN refers to the architecture used in
{[}\protect\hyperlink{ref-radford2015unsupervised}{14}{]}. For more
details, please see \protect\hyperlink{sec:netdetails}{Appendix A}.
}
\end{table}

\section{Experiments}\label{experiments}

\textbf{Datasets.} We apply this model to two video datasets: the
widely-used moving MNIST
{[}\protect\hyperlink{ref-srivastava2015unsupervised}{19}{]} and a new
dataset of real-world videos of 5th Avenue recorded from above, which
was collected by the authors. More details about these datasets,
including a sample frame from 5th Avenue, can be found in
\protect\hyperlink{sec:data-appendix}{Appendix C}.

\textbf{Baseline.} In each of these evaluations we compare with a model
which is identical to the factored model except that its transition
function \(p(Z_t | Z_{t-1})\) is not factored (referred to as the
\emph{entangled} model in our experiments). It can be viewed as a
special case of our factored model which has a single high-dimensional
latent factor, and its latent space has the same total dimension as the
corresponding factored model in each experiment. This comparison is
intended to be as tight as possible, with any differences between the
factored model and the baseline coming exclusively from the
factorization in the latent space.

\subsection{Evaluation}\label{evaluation}

\textbf{Lower bound.} We compare the variational lower bound achieved by
our factorized model with that of a non-factorized but otherwise
identical model. These experiments reveal the price in terms of data
fidelity that we pay for representing the data as independently-changing
factors. Table \ref{tbl:mi-elbo} shows that the factored model achieves
a lower bound on par with the entangled model.

\begin{figure}
\begin{floatrow}
\subfloat[Factored model]{
  \includegraphics[width=0.3\textwidth]{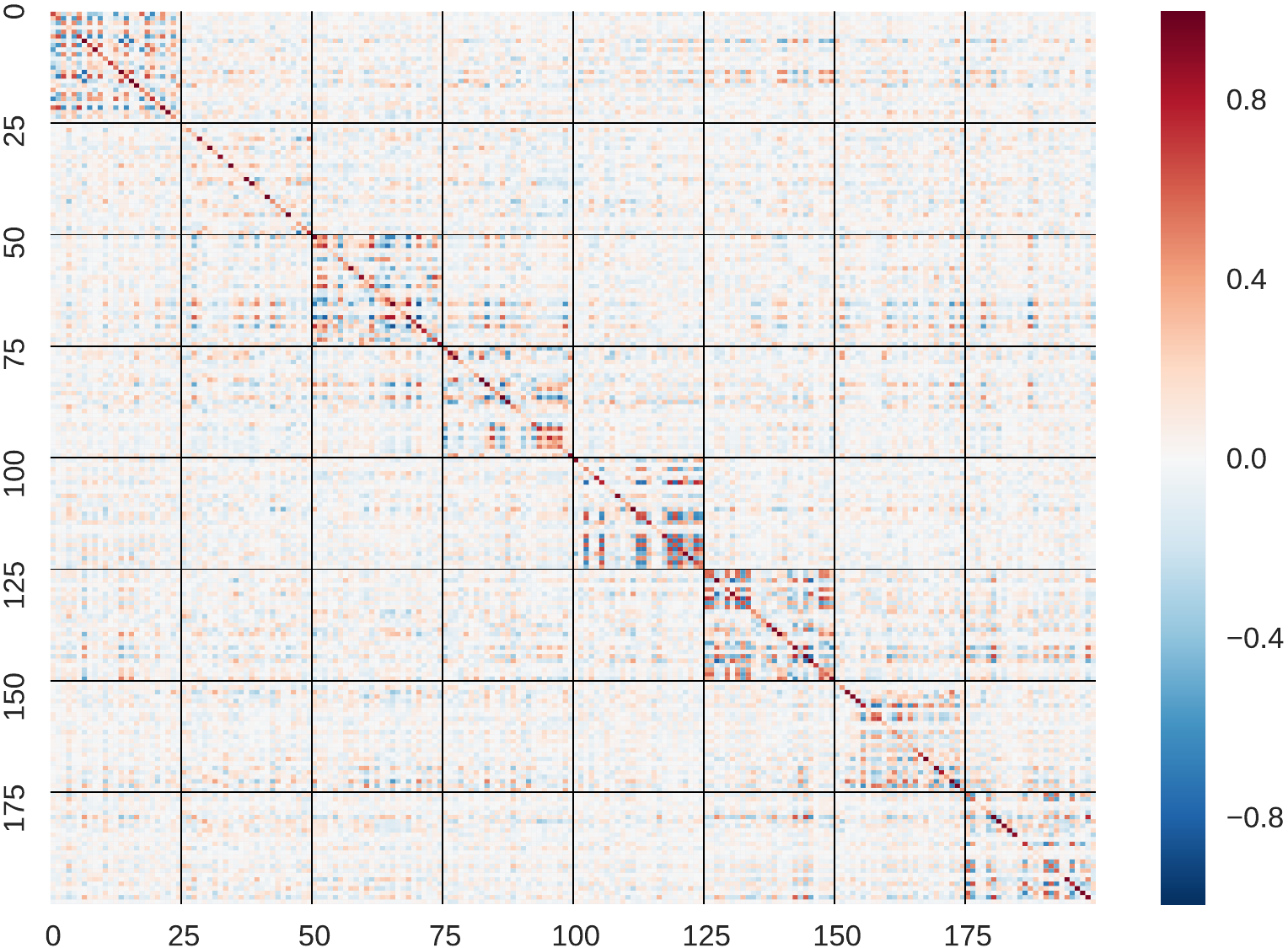}
}
\quad
\subfloat[Entangled model]{
  \includegraphics[width=0.3\textwidth]{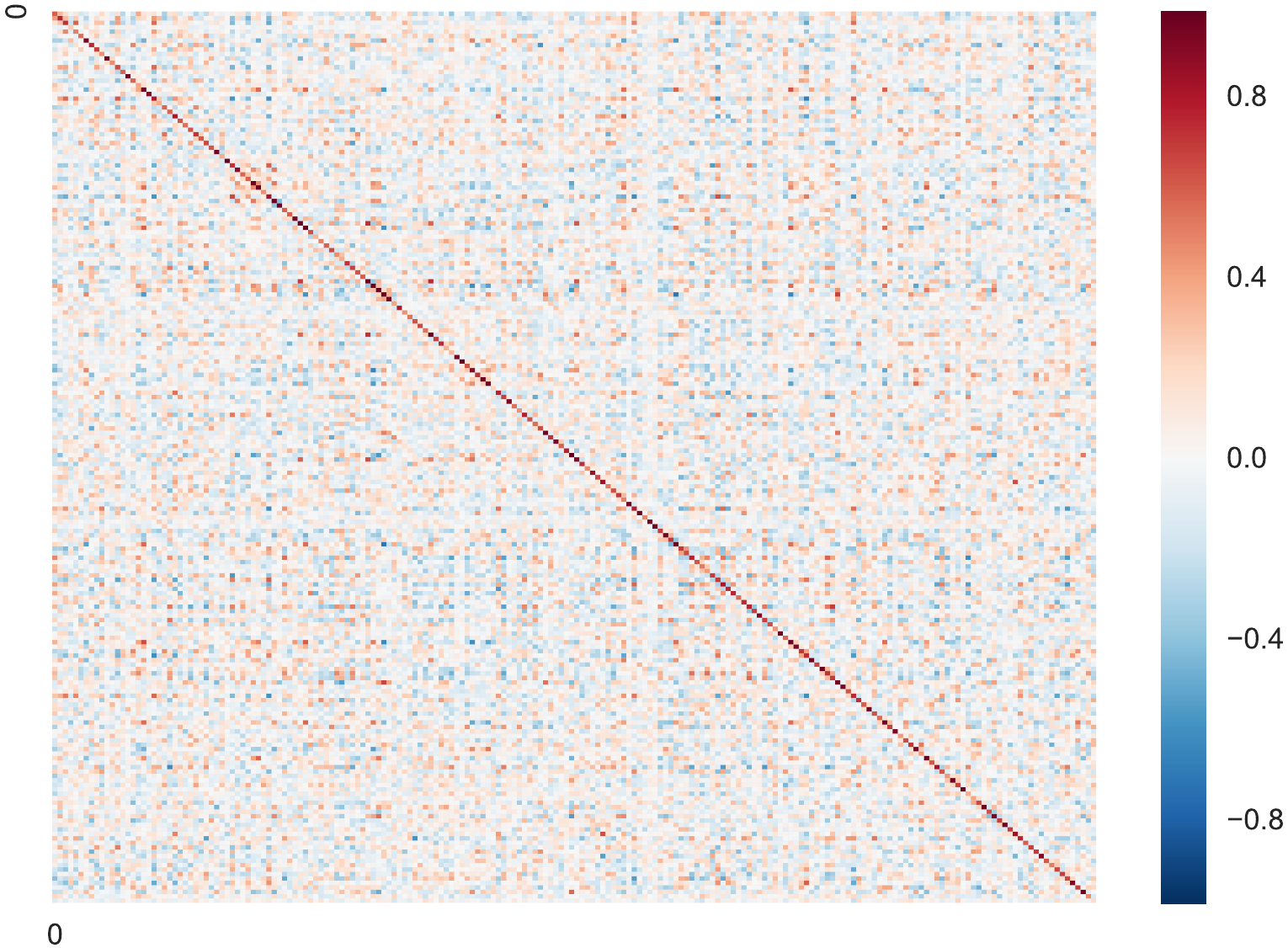}
}
\end{floatrow}

\caption{\label{fig:correlation-5thave}Heatmap of correlation between latent variables (shown as blocks) at time $t$ (y-axis) vs. at time $t+1$ (x-axis) on the 5th Avenue dataset. The block-diagonal structure exhibited here reflects that $z^i_{t+1}$ largely depends on its own previous state $z^i_{t}$ rather than on the other latent factors. The entangled model does not appear to have structure in its latent space. These plots were generated using models with two-layer transition networks; with deeper transitions there is almost zero linear dependence between any pair of latent units.}
\end{figure}

\textbf{Correlation structure.} By plotting the correlation between
samples from the approximate posterior over the latent variables at time
\(t\) and at time \(t+1\), we may observe whether our model has been
able to learn a representation for which \(z^i_{t+1}\) really does only
depend on \(z^i_{t}\) and not on the other latent factors. Note that
this only captures the \emph{linear} dependence of these variables; this
analysis helps to illustrate the structure of our latent space, but
should not be considered definitive. Figure \ref{fig:correlation-5thave}
shows that each latent factor is much more correlated with its own
previous state than with other latent factors.

\begin{table}[]
\centering
\begin{tabular}[c]{l l l r}
\toprule
dataset, model & \(\tilde{ \mathbb{I}}(z^i_{t+1}; z^i_{t})\) &
\(\tilde{\mathbb{I}}(z^i_{t+1}; z^{j \neq i}_{t})\) & ELBO\tabularnewline
\midrule
moving MNIST, two factors & 4.61 & 0.75 & -2902\tabularnewline
moving MNIST, entangled & 4.66 & 2.61 & -2896\tabularnewline
\midrule
5th Ave, eight factors & 2.26 & 0.28 & 6830\tabularnewline
5th Ave, entangled & 2.08 & 0.38 & 6800\tabularnewline
\bottomrule
\end{tabular}

\caption{\label{tbl:mi-elbo} Mutual information estimates and ELBO. In the factored
model, the previous value of the same latent variable is substantially
more predictive of its current value than are the previous values of any
other variables. This shows that our model is actually factorizing. See
Sec.~\ref{sec:mi} for details. The ELBO $\mathcal{L}(\theta, \phi; X)$
shows that our model pays a small or nonexistent price for factoring the
latent space into independently-evolving components.}
\end{table}

\textbf{Approximate mutual information.} \label{sec:mi} We use Kraskov's
method for estimating mutual information
{[}\protect\hyperlink{ref-kraskov2004estimating}{13}{]} to approximate
\(\mathbb{I}(z^i_{t+1}; z^i_{t})\) and compare it to
\(\mathbb{I}(z^i_{t+1}; z^{j \neq i}_{t})\). A latent factor \(z^i_{t}\)
should be much more informative about its own future \(z^i_{t+1}\) than
a different factor \(z^{j \neq i}_{t}\) is. The results, shown in Table
\ref{tbl:mi-elbo}, reveal that the evolution of each factor in the
factored models depends almost exclusively on their past.

For the entangled models, which do not have separate factors, there is
no \emph{a priori} subdivision into high- and low-mutual-information
segments of units. The reported scores were generated by creating 20
random partitionings of the latent units into \(k\) factors, then
reporting the mutual information numbers for the partitioning that had
the greatest difference between same-factor and cross-factor
information.

The entangled models show much more cross-factor information than the
factored models in our tests on moving MNIST, where we use two latent
factors. However, as the number of factors increases, the average
information between a pair of factors naturally diminishes. As a result
on 5th Avenue, where we subdivide the latent units into eight factors,
the entangled model shows cross-information almost as low as the
factored model.

\begin{figure}
\begin{floatrow}
\subfloat[Factorized reconstruction of moving MNIST]{
  \includegraphics[width=0.3\textwidth]{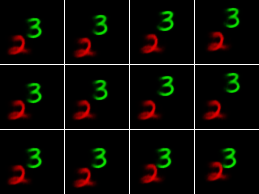}
}
\quad
\subfloat[Factorized reconstruction of 5th Avenue]{
  \includegraphics[width=0.3\textwidth]{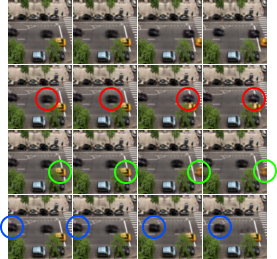}
}
\end{floatrow}

\caption{\label{fig:generations}\textbf{Left, top row:} reconstruction of the input sequence using all factors. \textbf{Left, middle row:} reconstructing the sequence varying only the first latent factor (of two). Only the green three moves, showing that factor 1 exactly represents the green digit. \textbf{Left, bottom row:} reconstructing the sequence varying only the second latent factor. Evolving factor two only affects the pose of the red digit. \textbf{Right, top row:} reconstruction of the input sequence using all latents. \textbf{Right, other rows:} reconstructions only varying a single latent factor. Several factors correspond to moving a single car in the image. Note that in the last row, the model removes the top-right black car; since the top-left car is moving forward quickly, it predicts that there must not be an obstacle in its way.}
\end{figure}

\textbf{Independent generations.} Finally, we evaluate qualitatively the
representations learned by our model. We infer the approximate posterior
\(q(Z | X)\), then set all of the latent variables fixed at their
\(t=2\) values given by \(q(Z_2 | x_{1,2})\). We then produce a sequence
of generations by picking a single variable \(z^i\) to vary, then for
each timestep \(t=2...n\) drawing a sample from
\(p(x | z^1_2 ... z^{i-1}_2, \boldsymbol{z^i_t}, z^{i+1}_2 ... z^{k}_2)\)
(note the single bolded factor \(\boldsymbol{z^i_t}\) varying with
\(t\)). That is, we hold all but one of the latent variables fixed and
allow the single one to vary with the posterior. This allows us to see
exactly what that single variable represents in this video. The images
generated by this process are shown in Figure \ref{fig:generations}.

\section{Discussion}\label{discussion}

By taking advantage of the structure present in video, our model can
pull apart latent factors which change independently and produce a
representation composed of semantically meaningful variables. The
approach is conceptually simple and based on the insight that if two
objects do not interact, they can be predicted independently. In the
future we hope to apply a richer family of approximations to scale to
more complex data.

\small

\section*{References}\label{references}
\addcontentsline{toc}{section}{References}

\hypertarget{refs}{}
\hypertarget{ref-bengio2013representation}{}
{[}1{]} Yoshua Bengio, Aaron Courville, and Pascal Vincent. 2013.
Representation learning: A review and new perspectives. \emph{Pattern
Analysis and Machine Intelligence, IEEE Transactions on} 35, 8 (2013),
1798--1828.

\hypertarget{ref-chang2016compositional}{}
{[}2{]} Michael B Chang, Tomer Ullman, Antonio Torralba, and Joshua B
Tenenbaum. 2016. A compositional object-based approach to learning
physical dynamics. \emph{arXiv preprint arXiv:1612.00341} (2016).

\hypertarget{ref-Denton17}{}
{[}3{]} Emily Denton and Vighnesh Birodkar. 2017. Unsupervised learning
of disentangled representations from video. \emph{CoRR} abs/1705.10915,
(2017).

\hypertarget{ref-grathwohl2016disentangling}{}
{[}4{]} Will Grathwohl and Aaron Wilson. 2016. Disentangling space and
time in video with hierarchical variational auto-encoders. \emph{arXiv
preprint arXiv:1612.04440} (2016).

\hypertarget{ref-hoffman2013stochastic}{}
{[}5{]} Matthew D Hoffman, David M Blei, Chong Wang, and John Paisley.
2013. Stochastic variational inference. \emph{The Journal of Machine
Learning Research} 14, 1 (2013), 1303--1347.

\hypertarget{ref-Hsu17}{}
{[}6{]} Wei-Ning Hsu, Yu Zhang, and James R. Glass. 2017. Unsupervised
learning of disentangled and interpretable representations from
sequential data. \emph{CoRR} abs/1709.07902, (2017).

\hypertarget{ref-Hyvarinen16}{}
{[}7{]} Aapo Hyvarinen and Hiroshi Morioka. 2016. Unsupervised feature
extraction by time-contrastive learning and nonlinear ica. In
\emph{NIPS}.

\hypertarget{ref-Janner17}{}
{[}8{]} Wu Janner M. 2017. Learning to generalize intrinsic images with
a structured disentangling autoencoder. In \emph{NIPS}.

\hypertarget{ref-kingma2014adam}{}
{[}9{]} Diederik Kingma and Jimmy Ba. 2014. Adam: A method for
stochastic optimization. \emph{arXiv preprint arXiv:1412.6980} (2014).

\hypertarget{ref-kingma2013auto}{}
{[}10{]} Diederik P Kingma and Max Welling. 2013. Auto-encoding
variational bayes. \emph{arXiv preprint arXiv:1312.6114} (2013).

\hypertarget{ref-klambauer2017self}{}
{[}11{]} Günter Klambauer, Thomas Unterthiner, Andreas Mayr, and Sepp
Hochreiter. 2017. Self-normalizing neural networks. \emph{arXiv preprint
arXiv:1706.02515} (2017).

\hypertarget{ref-Greff16}{}
{[}12{]} Mathias Berglund Klaus Greff Antti Rasmus and Juergen
Schmidhuber. 2016. Deep unsupervised perceptual grouping. In
\emph{NIPS}.

\hypertarget{ref-kraskov2004estimating}{}
{[}13{]} Alexander Kraskov, Harald Stögbauer, and Peter Grassberger.
2004. Estimating mutual information. \emph{Physical review E} 69, 6
(2004), 066138.

\hypertarget{ref-krishnan2017structured}{}
{[}14{]} Rahul G Krishnan, Uri Shalit, and David Sontag. 2017.
Structured inference networks for nonlinear state space models. In
\emph{AAAI}, 2101--2109.

\hypertarget{ref-kulkarni2015deep}{}
{[}15{]} Tejas D Kulkarni, William F Whitney, Pushmeet Kohli, and Josh
Tenenbaum. 2015. Deep convolutional inverse graphics network. In
\emph{Advances in neural information processing systems}, 2530--2538.

\hypertarget{ref-Fraccaro17}{}
{[}16{]} Ulrich Paquet Marco Fraccaro Simon Kamronn. 2017. A
disentangled recognition and nonlinear dynamics model for unsupervised
learning. In \emph{NIPS}.

\hypertarget{ref-Siddarth17}{}
{[}17{]} Jan-Willem Van de Meent N. Siddharth Brooks Paige. 2017.
Learning disentangled representations with semi-supervised deep
generative models. In \emph{NIPS}.

\hypertarget{ref-radford2015unsupervised}{}
{[}18{]} Alec Radford, Luke Metz, and Soumith Chintala. 2015.
Unsupervised representation learning with deep convolutional generative
adversarial networks. \emph{arXiv preprint arXiv:1511.06434} (2015).

\hypertarget{ref-srivastava2015unsupervised}{}
{[}19{]} Nitish Srivastava, Elman Mansimov, and Ruslan Salakhudinov.
2015. Unsupervised learning of video representations using lstms. In
\emph{International conference on machine learning}, 843--852.

\hypertarget{ref-Thomas17}{}
{[}20{]} Valentin Thomas, Jules Pondard, Emmanuel Bengio, Marc Sarfati,
Philippe Beaudoin, Marie-Jean Meurs, Joelle Pineau, Doina Precup, and
Yoshua Bengio. 2017. Independently controllable factors. \emph{arXiv
1708.01289} (2017).

\hypertarget{ref-whitney2016understanding}{}
{[}21{]} William F Whitney, Michael Chang, Tejas Kulkarni, and Joshua B
Tenenbaum. 2016. Understanding visual concepts with continuation
learning. \emph{arXiv preprint arXiv:1602.06822} (2016).

\hypertarget{ref-wiskott2002slow}{}
{[}22{]} Laurenz Wiskott and Terrence J Sejnowski. 2002. Slow feature
analysis: Unsupervised learning of invariances. \emph{Neural
computation} 14, 4 (2002), 715--770.

\newpage

\appendix

\normalsize

\section*{Appendix A: Network architecture
details}\label{sec:netdetails}
\addcontentsline{toc}{section}{Appendix A: Network architecture details}

All models are trained using the ADAM optimizer
{[}\protect\hyperlink{ref-kingma2014adam}{9}{]} with a learning rate of
\(3e{-4}\).

\subsection*{Inference network}\label{inference-network}
\addcontentsline{toc}{subsection}{Inference network}

For the inference network, which parameterizes the function
\(q(Z_t | Z_{t-1}, x_t)\), we use an architecture derived from the DCGAN
discriminator {[}\protect\hyperlink{ref-radford2015unsupervised}{18}{]}.
We use the discriminator architecture to encode the input image \(x\),
then add an additional input to take in the value of \(Z_{t-1}\). We
pass the inference network the \emph{predicted values}
\(p_\theta(Z_t|Z_{t-1})\) instead of having it do inference directly
from \(Z_{t-1}\). We found that this greatly sped up training as the
inference network doesn't have to learn the transition function in order
to fit to it. This is equivalent to sharing parameters between the
transition network and the inference network, though the transition
parameters are not updated here.

DCGAN image encoder:

\begin{Shaded}
\begin{Highlighting}[]
\NormalTok{Conv2d(}\DecValTok{3}\NormalTok{, }\DecValTok{64}\NormalTok{, kernel_size}\OperatorTok{=}\NormalTok{(}\DecValTok{4}\NormalTok{, }\DecValTok{4}\NormalTok{), stride}\OperatorTok{=}\NormalTok{(}\DecValTok{2}\NormalTok{, }\DecValTok{2}\NormalTok{), padding}\OperatorTok{=}\NormalTok{(}\DecValTok{1}\NormalTok{, }\DecValTok{1}\NormalTok{))}
\NormalTok{BatchNorm2d(}\DecValTok{64}\NormalTok{, eps}\OperatorTok{=}\FloatTok{1e-05}\NormalTok{, momentum}\OperatorTok{=}\FloatTok{0.1}\NormalTok{, affine}\OperatorTok{=}\VariableTok{True}\NormalTok{)}
\NormalTok{Conv2d(}\DecValTok{64}\NormalTok{, }\DecValTok{128}\NormalTok{, kernel_size}\OperatorTok{=}\NormalTok{(}\DecValTok{4}\NormalTok{, }\DecValTok{4}\NormalTok{), stride}\OperatorTok{=}\NormalTok{(}\DecValTok{2}\NormalTok{, }\DecValTok{2}\NormalTok{), padding}\OperatorTok{=}\NormalTok{(}\DecValTok{1}\NormalTok{, }\DecValTok{1}\NormalTok{))}
\NormalTok{BatchNorm2d(}\DecValTok{128}\NormalTok{, eps}\OperatorTok{=}\FloatTok{1e-05}\NormalTok{, momentum}\OperatorTok{=}\FloatTok{0.1}\NormalTok{, affine}\OperatorTok{=}\VariableTok{True}\NormalTok{)}
\NormalTok{Conv2d(}\DecValTok{128}\NormalTok{, }\DecValTok{256}\NormalTok{, kernel_size}\OperatorTok{=}\NormalTok{(}\DecValTok{4}\NormalTok{, }\DecValTok{4}\NormalTok{), stride}\OperatorTok{=}\NormalTok{(}\DecValTok{2}\NormalTok{, }\DecValTok{2}\NormalTok{), padding}\OperatorTok{=}\NormalTok{(}\DecValTok{1}\NormalTok{, }\DecValTok{1}\NormalTok{))}
\NormalTok{BatchNorm2d(}\DecValTok{256}\NormalTok{, eps}\OperatorTok{=}\FloatTok{1e-05}\NormalTok{, momentum}\OperatorTok{=}\FloatTok{0.1}\NormalTok{, affine}\OperatorTok{=}\VariableTok{True}\NormalTok{)}
\NormalTok{Conv2d(}\DecValTok{256}\NormalTok{, }\DecValTok{512}\NormalTok{, kernel_size}\OperatorTok{=}\NormalTok{(}\DecValTok{4}\NormalTok{, }\DecValTok{4}\NormalTok{), stride}\OperatorTok{=}\NormalTok{(}\DecValTok{2}\NormalTok{, }\DecValTok{2}\NormalTok{), padding}\OperatorTok{=}\NormalTok{(}\DecValTok{1}\NormalTok{, }\DecValTok{1}\NormalTok{))}
\NormalTok{BatchNorm2d(}\DecValTok{512}\NormalTok{, eps}\OperatorTok{=}\FloatTok{1e-05}\NormalTok{, momentum}\OperatorTok{=}\FloatTok{0.1}\NormalTok{, affine}\OperatorTok{=}\VariableTok{True}\NormalTok{)}
\NormalTok{Conv2d(}\DecValTok{512}\NormalTok{, }\DecValTok{32}\NormalTok{, kernel_size}\OperatorTok{=}\NormalTok{(}\DecValTok{4}\NormalTok{, }\DecValTok{4}\NormalTok{), stride}\OperatorTok{=}\NormalTok{(}\DecValTok{1}\NormalTok{, }\DecValTok{1}\NormalTok{))}
\end{Highlighting}
\end{Shaded}

Combining information about \(Z_{t-1}\) with information about \(x_t\):

\begin{Shaded}
\begin{Highlighting}[]
\NormalTok{transformed_x }\OperatorTok{=}\NormalTok{ Linear(}\DecValTok{32} \OperatorTok{->}\NormalTok{ z_dim)(encoder_output)}
\NormalTok{transformed_mu }\OperatorTok{=}\NormalTok{ Linear(z_dim }\OperatorTok{->}\NormalTok{ z_dim)( mu(Z_\{t\} }\OperatorTok{|}\NormalTok{ Z_\{t}\OperatorTok{-}\DecValTok{1}\NormalTok{\}) )}
\NormalTok{transformed_sigma }\OperatorTok{=}\NormalTok{ Linear(z_dim }\OperatorTok{->}\NormalTok{ z_dim)( sigma(Z_\{t\} }\OperatorTok{|}\NormalTok{ Z_\{t}\OperatorTok{-}\DecValTok{1}\NormalTok{\}) )}
\NormalTok{latent }\OperatorTok{=}\NormalTok{ Linear(z_dim }\OperatorTok{*} \DecValTok{3} \OperatorTok{->}\NormalTok{ z_dim)(transformed_x, transformed_mu, transformed_sigma)}
\NormalTok{output_mu }\OperatorTok{=}\NormalTok{ Linear(z_dim }\OperatorTok{->}\NormalTok{ z_dim)(latent)}
\NormalTok{output_sigma }\OperatorTok{=}\NormalTok{ Linear(z_dim }\OperatorTok{->}\NormalTok{ z_dim)(latent)}
\end{Highlighting}
\end{Shaded}

where \texttt{mu(Z\_\{t\}\ \textbar{}\ Z\_\{t-1\})} and
\texttt{sigma(Z\_\{t\}\ \textbar{}\ Z\_\{t-1\})} are the mean and
variance vectors respectively of the prediction
\(p_\theta(Z_t|Z_{t-1})\) and \texttt{z\_dim} is the number of latent
factors times the dimensionality of each factor. \texttt{output\_mu} and
\texttt{output\_sigma} are the mean and diagonal covariance of the
approximate posterior, and \texttt{output\_sigma} is actually output as
\(\log \sigma^2\) for numerical reasons. Each layer is followed by a
Leaky ReLU activation.

At \(t=0\), when there is no \(Z_{t-1}\), we pass all-zero vectors
instead.

\subsection*{Generator network}\label{generator-network}
\addcontentsline{toc}{subsection}{Generator network}

The generator network takes in a latent vector \(Z\) and produces a
pixelwise mean for an output image. It has this form:

\begin{Shaded}
\begin{Highlighting}[]
\NormalTok{Linear (z_dim }\OperatorTok{->}\NormalTok{ z_dim)}
\NormalTok{ConvTranspose2d(z_dim, }\DecValTok{512}\NormalTok{, kernel_size}\OperatorTok{=}\NormalTok{(}\DecValTok{4}\NormalTok{, }\DecValTok{4}\NormalTok{), stride}\OperatorTok{=}\NormalTok{(}\DecValTok{1}\NormalTok{, }\DecValTok{1}\NormalTok{))}
\NormalTok{BatchNorm2d(}\DecValTok{512}\NormalTok{, eps}\OperatorTok{=}\FloatTok{1e-05}\NormalTok{, momentum}\OperatorTok{=}\FloatTok{0.1}\NormalTok{, affine}\OperatorTok{=}\VariableTok{True}\NormalTok{)}
\NormalTok{ConvTranspose2d(}\DecValTok{512}\NormalTok{, }\DecValTok{256}\NormalTok{, kernel_size}\OperatorTok{=}\NormalTok{(}\DecValTok{4}\NormalTok{, }\DecValTok{4}\NormalTok{), stride}\OperatorTok{=}\NormalTok{(}\DecValTok{2}\NormalTok{, }\DecValTok{2}\NormalTok{), padding}\OperatorTok{=}\NormalTok{(}\DecValTok{1}\NormalTok{, }\DecValTok{1}\NormalTok{))}
\NormalTok{BatchNorm2d(}\DecValTok{256}\NormalTok{, eps}\OperatorTok{=}\FloatTok{1e-05}\NormalTok{, momentum}\OperatorTok{=}\FloatTok{0.1}\NormalTok{, affine}\OperatorTok{=}\VariableTok{True}\NormalTok{)}
\NormalTok{ConvTranspose2d(}\DecValTok{256}\NormalTok{, }\DecValTok{128}\NormalTok{, kernel_size}\OperatorTok{=}\NormalTok{(}\DecValTok{4}\NormalTok{, }\DecValTok{4}\NormalTok{), stride}\OperatorTok{=}\NormalTok{(}\DecValTok{2}\NormalTok{, }\DecValTok{2}\NormalTok{), padding}\OperatorTok{=}\NormalTok{(}\DecValTok{1}\NormalTok{, }\DecValTok{1}\NormalTok{))}
\NormalTok{BatchNorm2d(}\DecValTok{128}\NormalTok{, eps}\OperatorTok{=}\FloatTok{1e-05}\NormalTok{, momentum}\OperatorTok{=}\FloatTok{0.1}\NormalTok{, affine}\OperatorTok{=}\VariableTok{True}\NormalTok{)}
\NormalTok{ConvTranspose2d(}\DecValTok{128}\NormalTok{, }\DecValTok{64}\NormalTok{, kernel_size}\OperatorTok{=}\NormalTok{(}\DecValTok{4}\NormalTok{, }\DecValTok{4}\NormalTok{), stride}\OperatorTok{=}\NormalTok{(}\DecValTok{2}\NormalTok{, }\DecValTok{2}\NormalTok{), padding}\OperatorTok{=}\NormalTok{(}\DecValTok{1}\NormalTok{, }\DecValTok{1}\NormalTok{))}
\NormalTok{BatchNorm2d(}\DecValTok{64}\NormalTok{, eps}\OperatorTok{=}\FloatTok{1e-05}\NormalTok{, momentum}\OperatorTok{=}\FloatTok{0.1}\NormalTok{, affine}\OperatorTok{=}\VariableTok{True}\NormalTok{)}
\NormalTok{ConvTranspose2d(}\DecValTok{64}\NormalTok{, }\DecValTok{3}\NormalTok{, kernel_size}\OperatorTok{=}\NormalTok{(}\DecValTok{4}\NormalTok{, }\DecValTok{4}\NormalTok{), stride}\OperatorTok{=}\NormalTok{(}\DecValTok{2}\NormalTok{, }\DecValTok{2}\NormalTok{), padding}\OperatorTok{=}\NormalTok{(}\DecValTok{1}\NormalTok{, }\DecValTok{1}\NormalTok{))}
\end{Highlighting}
\end{Shaded}

where \texttt{z\_dim} is the number of latent factors times the
dimensionality of each factor. Each layer is followed by a ReLU
activation. We use a fixed variance in our Normal observation model of
0.25 for moving MNIST and 0.05 for 5th Avenue. This variance
hyperparameter may be tuned to balance the tradeoff between fitting the
predictions and making tight reconstructions.

\subsection*{Transition network}\label{transition-network}
\addcontentsline{toc}{subsection}{Transition network}

Each latent factor has its own transition function
\(p_\theta(z^i_t | z^i_{t-1})\). Each of these has the following form:

\begin{Shaded}
\begin{Highlighting}[]
\NormalTok{Linear (latent_dim }\OperatorTok{->} \DecValTok{128}\NormalTok{)}
\NormalTok{Linear (}\DecValTok{128} \OperatorTok{->} \DecValTok{128}\NormalTok{)}
\NormalTok{Linear (}\DecValTok{128} \OperatorTok{->} \DecValTok{128}\NormalTok{)}
\NormalTok{Linear (}\DecValTok{128} \OperatorTok{->} \DecValTok{128}\NormalTok{)}
\NormalTok{Linear (}\DecValTok{128} \OperatorTok{->}\NormalTok{ latent_dim }\OperatorTok{*} \DecValTok{2}\NormalTok{)}
\end{Highlighting}
\end{Shaded}

where \texttt{latent\_dim} is the dimensionality of a single latent
factor and each layer is followed by a SELU activation
{[}\protect\hyperlink{ref-klambauer2017self}{11}{]}. The
\texttt{latent\_dim\ *\ 2} output is for the mean and (diagonal)
variance vectors for the Normal distribution. The variance is produced
as \(\log \sigma^2\) for numerical reasons.

\hypertarget{sec:elbo-derivation}{\section*{Appendix B: Deriving the
ELBO}\label{sec:elbo-derivation}}
\addcontentsline{toc}{section}{Appendix B: Deriving the ELBO}

For this section the factored form of the transitions
\(p_\theta(Z_t | Z_{t-1}) = \prod_{i=1}^k p_\theta(z^i_t | z^i_{t-1})\)
is not relevant. As such our development here will use the more general
non-factored form, and we can substitute in our factored special case
later. For simplicity of notation we will use \(p(\cdot)\) in place of
\(p_\theta(\cdot)\). Likewise we use \(q(\cdot)\) in place of
\(q_\phi(\cdot)\) to represent our variational approximation function
with parameters \(\phi\).

We begin with the form of our latent-variable generative model.

\begin{align*}
\log p(X) &= \log \int p(X|Z) p(Z) dZ \\
&= \log \int p(X|Z) p(Z_1) p(Z_2 | Z_1) ... p(Z_{t_{max}} | Z_{t_{max}-1}...Z_1) dZ \\
\end{align*}

Since the series \(Z\) is Markov,

\begin{align*}
= \log \int p(X|Z) p(Z_1) p(Z_2 | Z_1) ... p(Z_{t_{max}} | Z_{t_{max}-1}) dZ \\
\end{align*}

We introduce our variational auxiliary functions:

\begin{align*}
&= \log \int p(X|Z) p(Z_1) p(Z_2 | Z_1) ... p(Z_{t_{max}} | Z_{t_{max}-1}) \frac{q(Z_1 | X) q(Z_2 | Z_1, X) ... q(Z_{t_{max}} | Z_{t_{max}-1}, X)}{q(Z_1 | X) q(Z_2 | Z_1, X) ... q(Z_{t_{max}} | Z_{t_{max}-1}, X})) dZ \\
&= \log \int p(X|Z) \frac{p(Z_1)}{q(Z_1 | X)} \frac{p(Z_2 | Z_1)}{q(Z_2 | Z_1, X)} ... \frac{p(Z_{t_{max}} | Z_{t_{max}-1})}{q(Z_{t_{max}} | Z_{t_{max}-1}, X)} q(Z_1 | X) q(Z_2 | Z_1, X) ... q(Z_{t_{max}}) dZ \\
\end{align*}

We may now convert this integral to an expectation with respect to
\(q(Z|X) = q(Z_1 | X) q(Z_2 | Z_1, X) ... q(Z_{t_{max}} | Z_{t_{max}-1}, X)\):

\begin{align*}
= \log \expect_{Z \sim q(Z | X)} p(X|Z) \frac{p(Z_1)}{q(Z_1 | X)} \frac{p(Z_2 | Z_1)}{q(Z_2 | Z_1, X)} ... \frac{p(Z_{t_{max}} | Z_{t_{max}-1})}{q(Z_{t_{max}} | Z_{t_{max}-1}, X)}
\end{align*}

By Jensen's inequality,

\begin{align*}
&>= \expect_{Z \sim q(Z | X)} \log \left\{p(X|Z) \frac{p(Z_1)}{q(Z_1 | X)} \frac{p(Z_2 | Z_1)}{q(Z_2 | Z_1, X)} ... \frac{p(Z_{t_{max}} | Z_{t_{max}-1})}{q(Z_{t_{max}} | Z_{t_{max}-1}, X)} \right\} \\
&= \expect_{Z \sim q(Z | X)} \left[\
\log p(X|Z) + \
\log \frac{p(Z_1)}{q(Z_1 | X)} + \
\log \frac{p(Z_2 | Z_1)}{q(Z_2 | Z_1, X)} + ... + \
\log \frac{p(Z_{t_{max}} | Z_{t_{max}-1})}{q(Z_{t_{max}} | Z_{t_{max}-1}, X)} \right] \\
\begin{split}
= \expect_{Z \sim q(Z | X)}  \log p(X|Z) \
- \expect_{Z_1 \sim q(Z_1 | X)}  \
\log \frac{q(Z_1 | X)}{p(Z_1)} \
- \expect_{ \substack{ Z_1 \sim q(Z_1 | X) \\ Z_2 \sim q(Z_2 | Z_1, X) }} \
\log \frac{q(Z_2 | Z_1, X)}{p(Z_2 | Z_1)} \
- ... \\
- \expect_{\substack{ Z_{t_{max}-1} \sim q(Z_{t_{max}-1} | Z_{t_{max}-2}, X) \\ Z_{t_{max}} \sim q(Z_{t_{max}} | Z_{t_{max}-1}, X) }} \
\log \frac{q(Z_{t_{max}} | Z_{t_{max}-1}, X)}{p(Z_{t_{max}} | Z_{t_{max}-1})}
\end{split}
\end{align*}

Realizing that the expectations \(\mathbb{E} \log \frac{q}{p}\) are KL
divergences gives us an objective we can optimize:

\begin{align*}
\mathcal{L}(\theta, \phi; X) \
=& \expect_{Z \sim q_\phi(Z | X)}  \log p_\theta(X|Z) \\
&- D_{KL}(q_\phi(Z_1 | X) || p_\theta(Z_1)) \
- \sum_{t=2}^{t_{max}} \expect_{Z_{t-1} \sim q(Z_{t-1} | Z_{t-2}, X) } \
D_{KL}(q_\phi(Z_t | Z_{t-1}, X) || p_\theta(Z_{t} | Z_{t-1})) \\
\le& \log p_\theta(X) \nonumber
\end{align*}

This lower bound lies below the true log-probability by an additive term
of \(D_{KL}(q(Z | X) || p(Z | X))\)
{[}\protect\hyperlink{ref-kingma2013auto}{10}{]}. As the variational
approximation \(q(Z | X)\) improves (i.e., approaches the true
posterior), this lower bound approaches the true log-likelihood of the
data.

This bound would hold for any function \(p_\theta(Z_{t} | Z_{t-1})\).
Our model factors this general transition function:

\[ p_\theta(Z_{t} | Z_{t-1}) = \prod_{i=1}^k p_\theta(z^i_{t} | z^i_{t-1}) \]

which corresponds to a hidden Markov model with multiple Markov chains
running in parallel in the latent space.

\hypertarget{sec:data-appendix}{\section*{Appendix C: Details on
datasets}\label{sec:data-appendix}}
\addcontentsline{toc}{section}{Appendix C: Details on datasets}

\subsection*{Moving MNIST}\label{moving-mnist}
\addcontentsline{toc}{subsection}{Moving MNIST}

This dataset consists of two digits from the MNIST dataset bouncing in a
64x64 pixel frame. Each digit is on a separate plane of the input (i.e.,
one is red and the other is green). The digits have randomized starting
location and velocity vector for each sequence, but their motion is
deterministic over the course of the sequence and the digits do not
interact.

\subsection*{5th Avenue}\label{th-avenue}
\addcontentsline{toc}{subsection}{5th Avenue}

\begin{figure}
\centering
\includegraphics[width=0.80000\textwidth]{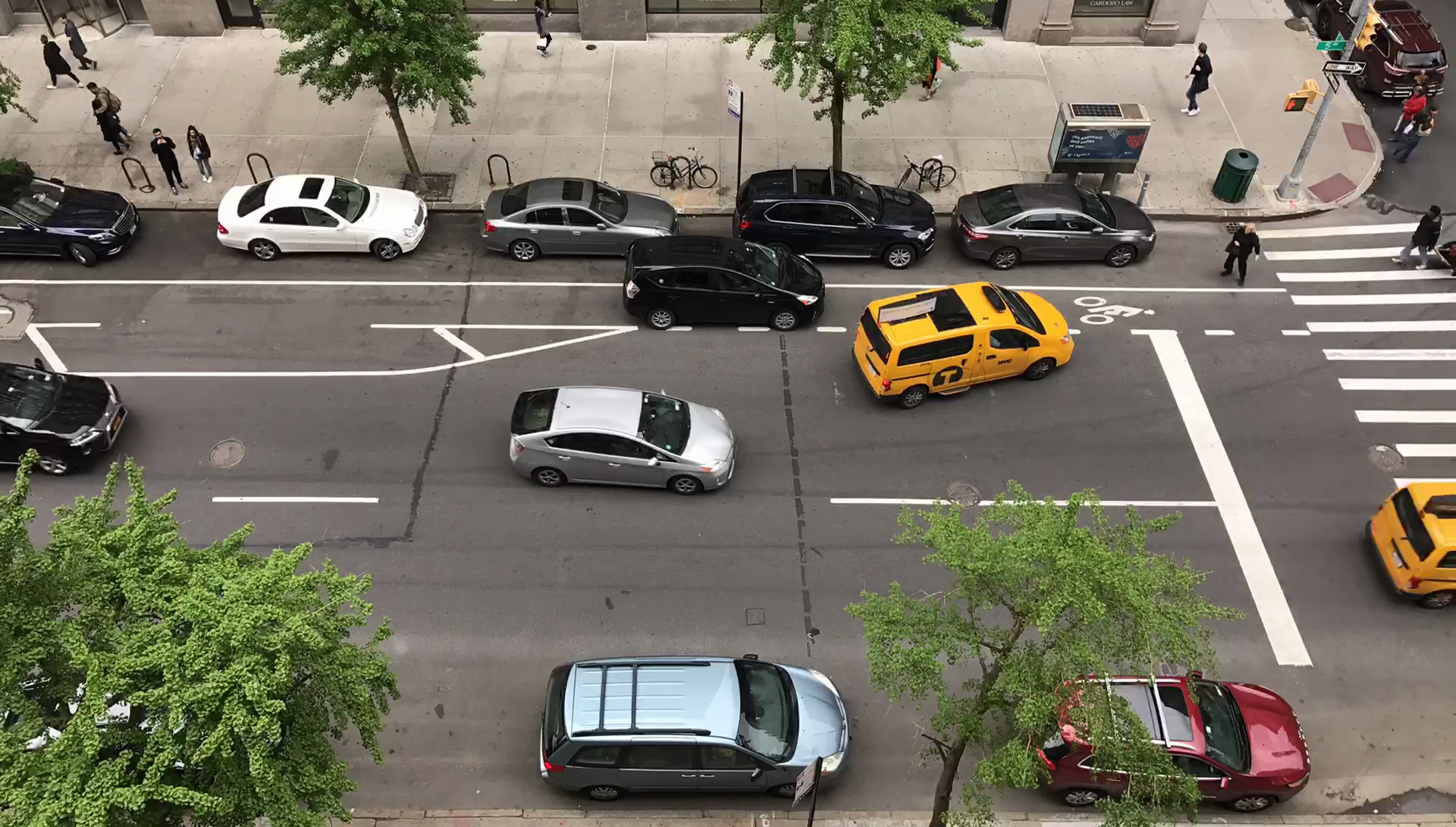}
\caption{An example image from the collected 5th Avenue dataset of urban
videos. These videos include much of the complexity of the real world,
including textures, lighting, and uncertain dynamics, while remaining
simple enough to model for several frames with a neural
network.}\label{fig:5th-ave}
\end{figure}

The 5th Avenue dataset has greater complexity in its visuals and its
dynamics than moving MNIST, but was designed to be simple enough to
model with some fidelity using contemporary techniques. It consists of
around 20 hours of video sampled at 2 frames per second. The videos were
recorded from the 5th floor of a building overlooking 5th Avenue in
Manhattan and show the busy street scene below including pedestrians and
passing cars. Each video was recorded with a fixed camera position;
between videos the camera position is nearly the same but may vary
slightly. The data includes global variations such as time of day and
weather. It was recorded on an iPhone 7 at 1080p resolution, though in
our experiments we resize it to 64x64. A representative example image is
shown in Figure \ref{fig:5th-ave}.

\end{document}